\documentclass[11pt,a4paper]{article}
\usepackage{times,latexsym}
\usepackage{url}
\usepackage[T1]{fontenc}
\usepackage[utf8]{inputenc} 
\usepackage[T1]{fontenc}    
\usepackage{hyperref}       
\usepackage{url}            
\usepackage{booktabs}       
\usepackage{amsfonts}       
\usepackage{nicefrac}       
\usepackage{makecell}
\usepackage[protrusion=false]{microtype}      
\microtypesetup{protrusion=false}
\usepackage{xcolor}         
\usepackage[table]{xcolor}
\usepackage{tabularx}
\usepackage{multirow}
\usepackage{multicol}
\usepackage{graphicx}
\usepackage{amsmath}
\usepackage{arydshln}
\usepackage{subfigure}
\usepackage{wrapfig,lipsum,booktabs}
\usepackage{color}
\usepackage[most]{tcolorbox}
\usepackage{enumitem}
\usepackage{algorithm}
\usepackage{algorithmic}
\usepackage{amssymb}

\usepackage[acceptedWithA]{tacl2021v1}

\usepackage{xspace,mfirstuc,tabulary}

\newif\iftaclinstructions
\taclinstructionsfalse 
\iftaclinstructions
\renewcommand{\confidential}{}
\renewcommand{\anonsubtext}{(No author info supplied here, for consistency with
TACL-submission anonymization requirements)}
\newcommand{\instr}
\fi

\iftaclpubformat 

\else

\fi

\title{\textsc{Homer}: Understanding Long-form Videos with Hierarchical Memory and Agentic Reasoning}

\author{
  Yixin Ji$^{1}$, 
  Fanghua Ye$^{2}$,
  Juntao Li$^{1}$\thanks{\; Corresponding authors.}, 
  Bo Zhao$^{2}$,
  Zexuan Qiu$^{2}$,\\
  \textbf{Zhaopeng Tu}$^{2}$\footnotemark[1],
  \textbf{Liefeng Bo}$^{2}$,
  \textbf{Min Zhang}$^{1}$ \\
  $^{1}$Soochow University \\
  $^{2}$Tencent Hunyuan Multimodal Department\\
  \texttt{\{jiyixin169, fanghua.ye.21\}@gmail.com} \\
  \texttt{\{ljt, minzhang\}@suda.edu.cn}
}

\date{}

\begin{document}
\maketitle
\begin{abstract}
Multimodal large language models excel on short clips but struggle on hour-long videos in an online setting, where frames are processed incrementally under limited memory.
Existing online methods either retain compact visual representations that lack semantic structure, or build higher-level memory stores organized around temporal proximity rather than explicit causal links, leaving multi-hop narrative reasoning to be reconstructed by the LLM at every query.
We bridge this gap with \textsc{Homer}, a Hierarchical Online Memory Exploration and Reasoning framework.
\textsc{Homer}'s memory mirrors the multi-scale structure of long videos, ranging from raw perception, to recurring entities, to events connected by explicit temporal and causal relations.
Its agentic reasoner then explores this memory the way humans do, locating the relevant scene, looking up details, and composing the answer through multi-round memory retrieval, with a harness that verifies and corrects each step.
\textsc{Homer} outperforms the previous best agent method by $+5.5$, $+10.8$, and $+4.4$ points on M3-Bench-robot, M3-Bench-web, and Video-MME-Long, and consistently lifts three various LLM backbones, indicating a model-agnostic structural capability for grounded retrieval over long videos.\footnote{Code is available at \url{https://github.com/Dereck0602/Homer}}

\end{abstract}

\section{Introduction}
Reasoning over long-form videos is a basic requirement for multimodal large language models (MLLMs) used in embodied intelligence~\citep{shen2026videovla,pai2025mimic}, autonomous agents~\citep{chen2025gui}, and world modeling~\citep{ren2025videoworld}, yet MLLMs that handle short clips fluently still struggle on such long-form content~\citep{zou2024seconds,bai2025qwen3}.
Videos spanning tens of minutes to hours contain thousands of visually redundant frames, complex narrative structures, and dense interactions among characters.
The resulting token count far exceeds the context window of current models, and even with sufficient context length, critical details are dispersed across the timeline, making it difficult for attention mechanisms to locate and integrate relevant evidence~\citep{tang2025revisiting,dou2026cl}.
Long-video understanding therefore demands both compressing redundancy to fit the model's context and pinpointing fine-grained evidence dispersed across that context.

Existing approaches address these demands from different angles.
To reduce the token count, sparse frame selection methods~\citep{wang2025videotree,tang2025aks} subsample a small set of representative frames, trading temporal coverage for efficiency, while token compression methods~\citep{li2024llamavid,shen2024longvu} reduce the number of visual tokens per frame so that more frames fit within the context window, though aggressive compression may discard fine-grained details.
To improve evidence localization, agentic reasoning methods~\citep{wang2024videoagent,shen2025vgent} equip MLLMs with visual tools to iteratively manipulate and reason over videos, achieving stronger multi-hop reasoning performance.
Most of these methods, however, assume an offline setting where the full video is available at query time and segments can be re-retrieved on demand when the first evidence is insufficient.


Many real-world applications instead require an online setting where frames arrive as a continuous stream and storing the full video is infeasible, demanding to incrementally accumulate and organize information as the video progresses.
Prior online memory approaches adopt visual token compression~\citep{wang2025videollamb,xie2026fluxmem} or textual memory banks~\citep{song2024moviechat,he2024malmm,long2026seeing} to maintain memory-efficient representations.
While these methods successfully control memory size, they face a fundamental tension between information coverage and retrieval difficulty.
Coarse-grained memories are easy to search but discard fine details, while fine-grained memories preserve detail but become hard to navigate at hour scales.
Moreover, none of these methods explicitly model the temporal and causal relationships between events, making multi-hop narrative reasoning particularly challenging.

The limitations above motivate two principles for online long-video understanding. First, the memory should separate broad narrative coverage from fine-grained perceptual detail and treat the temporal and causal relations between events as first-class, rather than leave multi-hop reasoning to be reconstructed by the LLM at every query. Second, the reasoning loop should be able to detect when the first retrieval is insufficient and recover, instead of committing to a single retrieval pass that cannot self-correct. We address both with \textsc{Homer}, a Hierarchical Online Memory Exploration and Reasoning framework that pairs a hierarchical memory with an agentic reasoner. The memory consists of three complementary layers, namely a \emph{Perceptual Buffer} retaining visual keyframes for revisit, an \emph{EntityGraph} organizing episodic and semantic memories around identified characters, and an \emph{EventGraph} adding explicit temporal and causal edges between event segments to support multi-hop narrative reasoning. On top of this memory, the agentic reasoner decomposes each question into focused subtasks and runs multi-round retrieval supervised by a harness that verifies and corrects each step, while a self-evolution mechanism accumulates reusable retrieval skills across questions.

We summarize our contributions as follows.
\begin{itemize}[leftmargin=*,itemsep=2pt,topsep=2pt]
    \item We introduce \textsc{Homer}, an online long-video understanding framework whose hierarchical memory separates narrative coverage from perceptual detail and adds explicit temporal-causal edges, addressing both the coverage-retrieval tension and the absence of structured event reasoning in prior online memories.
    \item We design an agentic reasoning loop driven by a verify-and-correct harness and a self-evolution mechanism, letting the agent recover from incorrect retrievals and accumulate reusable skills across questions.
    \item Across M3-Bench-robot, M3-Bench-web, and Video-MME-Long, \textsc{Homer} outperforms the previous best agent method by $+5.5$, $+10.8$, and $+4.4$ points. The gain is structural and model-agnostic, letting an open-source agentic backbone match a closed-source multimodal model without additional training.
\end{itemize}

\section{Related Work}
Existing approaches to long-video reasoning can be broadly divided along the processing paradigm axis into \textbf{offline} methods that access the entire video before answering and \textbf{online} methods that process frames incrementally under causality and memory constraints.

\paragraph{Offline Long-Video Reasoning.}
Offline methods either compress the visual input or invoke tools on demand.
\emph{Sparse frame selection} reduces the input to a small set of representative frames. For example, VideoTree~\citep{wang2025videotree} progressively refines a hierarchical tree of frames conditioned on the query, and AKS~\citep{tang2025aks} formulates key frame selection as a budget-constrained optimization that maximizes visual-textual relevance. FOCUS~\citep{zhu2025focus}, FocusGraph~\citep{zemskova2026focusgraph} and LongVLM~\citep{weng2024longvlm} extend this idea with region-level scoring, graph-structured dependencies, and multi-granularity aggregation.
\emph{Agentic reasoning} instead equips an MLLM with tool-use and planning to iteratively retrieve relevant segments. VideoAgent~\citep{wang2024videoagent} treats the LLM as a central agent that aggregates evidence through VLM and CLIP tools, while VideoMind~\citep{liu2026videomind} introduces role-based Planner-Grounder-Verifier-Answerer reasoning. DrVideo~\citep{ma2025drvideo}, LongVT~\citep{longvt2026}, Vgent~\citep{shen2025vgent} and VLog~\citep{lin2025vlog} further explore document-style retrieval, native tool calling, graph-based RAG, and narration-vocabulary indexing. With full-video access, these methods can reduce the context through sampling, indexing, or retrieval, and then return to the source video to localize fine-grained evidence, thereby balancing context reduction and evidence localization in the offline setting.

\begin{figure*}[t]
\centering
\includegraphics[width=\textwidth]{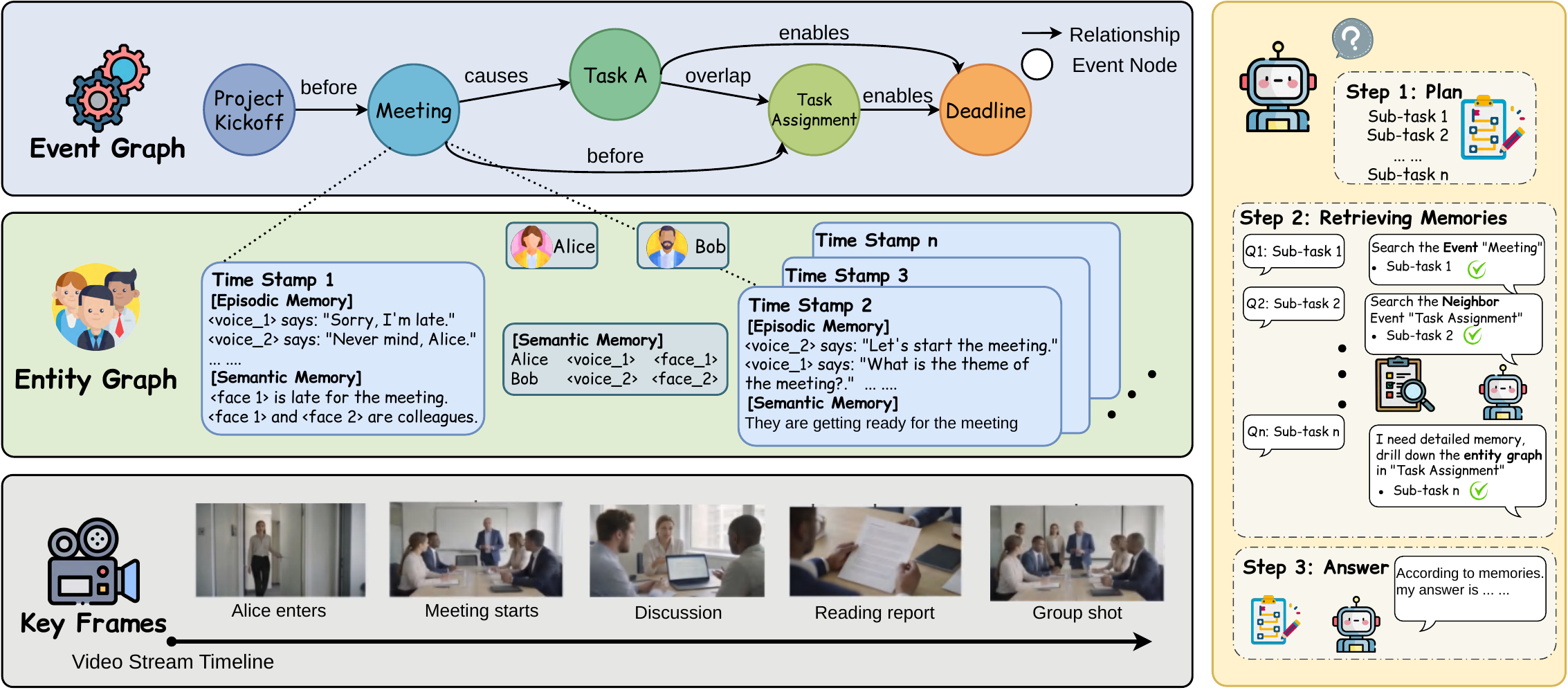}
\vspace{-20pt} 
\caption{Overview of \textsc{Homer}. Each incoming clip is processed online into a \emph{Perceptual Buffer} of key frames, an \emph{Entity Graph} of character-centric episodic and semantic memories, and an \emph{Event Graph} of narrative segments linked by temporal and causal edges. At question time, an agentic reasoner decomposes the query and performs multi-turn retrievals at progressively finer granularity to produce the final answer.}
\label{fig:framework}
\end{figure*}

\paragraph{Online Long-Video Reasoning.}
In streaming scenarios, frames arrive sequentially and the model cannot freely rewind to arbitrary past content, so the online long-video reasoning methods compress history into limited storage for evidence location.
One line of work retains compact visual tokens. LLaMA-VID~\citep{li2024llamavid} represents each frame with two tokens, while LongVU~\citep{shen2024longvu} and FluxMem~\citep{xie2026fluxmem} adaptively remove spatiotemporal redundancy. StreamingTOM~\citep{chen2025streamingtom} and StreamMem~\citep{yang2025streammem} further bound online computation through causal temporal reduction, KV-cache merging, and query-agnostic KV memory. VideoLLaMB~\citep{wang2025videollamb} uses recurrent memory bridges to encode long streaming videos with bounded memory. These methods achieve strong compression and can preserve fine-grained visual cues, but offer limited semantic structure for navigating evidence across distant segments.
Another line builds higher-level memory stores over the video stream. MovieChat~\citep{song2024moviechat} converts dense visual tokens into sparse short- and long-term memories, MA-LMM~\citep{he2024malmm} stores past video information in an online memory bank, and Flash-VStream~\citep{zhang2024flashvstream} combines a low-capacity context memory for temporal abstraction with a high-capacity augmentation memory for spatial detail retrieval. StreamForest~\citep{streamforest2025} further organizes history as an event-level memory forest guided by temporal distance and content similarity. 
Recent work begins to combine richer memory with agentic control. M3-Agent~\citep{long2026seeing} organizes an entity-centric multimodal memory around face and voice anchors and performs multi-round retrieval, yet its graph lacks explicit event-level causal links. MM-Mem~\citep{lian2026verbatim} introduces a pyramidal memory grounded in Fuzzy-Trace Theory and optimizes compression via a Semantic Information Bottleneck, but relies on a fixed retrieval strategy without adaptive planning.
Concurrent work PyraVid~\citep{yan2026pyravid} also constructs a hierarchical memory pyramid, but aggregates evidence through automatic graph expansion and pruning. Our framework differs by employing an agent that actively plans retrieval at each step and accumulates reusable reasoning skills across questions via self-evolution.
Our work follows this online memory-based paradigm, but organizes streaming video history as a hierarchical memory with explicit temporal and causal links between events. This design brings the multi-hop evidence localization ability of offline agentic reasoning into an online setting, where reasoning must rely on incrementally constructed and limited memory.

\section{Method}

We introduce \textsc{Homer} (Hierarchical Online Memory Exploration and Reasoning), a framework for online long video understanding inspired by how humans perceive and reason about extended temporal experiences. 
Cognitive science suggests that human memory operates at multiple levels: a perceptual level that retains raw visual impressions, an entity level where episodic and semantic memories are organized around specific individuals, and an event level that links episodes into narrative structures~\citep{tulving1972episodic,zacks2007event,hahamy2023human}.
In this section, we model three complementary memory components (Figure~\ref{fig:framework}):
\begin{itemize}[nosep,leftmargin=*]
    \item A \emph{Perceptual Buffer} $\mathcal{P}$ that retains visual keyframes as raw visual snapshots of each clip;
    \item An \emph{Entity Graph} $\mathcal{G}_\text{Ent}$ that organizes entity-level episodic and semantic memories around identified characters;
    \item An \emph{Event Graph} $\mathcal{G}_E$ that captures the high-level narrative structure through event segments linked by temporal and causal relations.
\end{itemize}
An agentic control process then navigates this three-layer memory through multi-turn retrieval and reasoning to answer questions.

\subsection{Perceptual Buffer: Key Frame Extraction}
\label{sec:perceptual}

Given a continuous video stream segmented into fixed-length clips $\{c_1, c_2, \ldots \}$, the Perceptual Buffer retains a small set of representative key frames per clip as visual snapshots. 
Analogous to the Perceptual Representation System~\citep{schacter1990perceptual}, these key frames provide raw visual evidence that can be revisited during reasoning when high-level memories alone are insufficient.

\noindent{\textbf{Shot Boundary Detection.}}
A video clip often spans several camera shots with visually distinct content. To ensure that the selected key frames cover this visual diversity, we segment each clip into shots and extract at least one key frame per shot. We adopt a lightweight frame-differencing approach for shot boundary detection. During a single decoding pass, the detector computes the mean absolute pixel intensity difference between consecutive sampled grayscale frames. Let $I_i$ denote the grayscale image at sample index $i$. A hard cut is declared at frame $i$ when
\begin{equation}
    d_i = \frac{1}{HW}\sum_{x,y} |I_i(x,y) - I_{i-1}(x,y)| > \tau_{\text{scene}},
\end{equation}
where $H$ and $W$ are the height and width of the frame, and $\tau_{\text{scene}}$ is a fixed threshold. This partitions the clip into shots $\{S_1, S_2, \ldots, S_K\}$.

\noindent{\textbf{Quality Score and Selection.}}
Within each shot, we select the frame that best represents the visual content. An uniform sampling strategy often yields blurry transitional frames or uninformative static frames (e.g., black screens, title cards). We therefore design a \emph{sharpness-entropy joint quality score} that combines two complementary signals. Sharpness is measured by the Laplacian variance $\text{Sharp}(f) = \text{Var}(\nabla^2 I_f)$, which yields high values for frames rich in edges and fine details but low values for out-of-focus or motion-blurred frames. Color entropy $H(f) = -\sum_{b=1}^{B} h_b \log_2 h_b$ quantifies color diversity through the information entropy of the HSV hue histogram, where $h_b$ is the normalized frequency of the $b$-th bin. Frames dominated by a single color (solid backgrounds, text overlays) yield near-zero entropy and are effectively down-weighted. The two signals are combined as:
\begin{equation}
    Q(f) = \text{Sharp}(f) \cdot \bigl(1 + w \cdot H(f)\bigr),
    \label{eq:quality}
\end{equation}
where $w$ controls the relative importance of color diversity.
Key frames are then selected in two stages to balance diversity and quality. In the first stage, the highest-scoring frame from each shot is selected to ensure that every visually distinct segment is represented. If the target number of key frames $n$ exceeds the number of shots, the second stage fills remaining slots by greedily selecting the globally highest-scoring frames from the candidate pool, subject to a minimum temporal separation $\delta_{\min}$ that prevents redundant nearby frames. The resulting key frames $\mathcal{P}_t = \{p_1, p_2, \ldots, p_n\}$ are indexed by clip ID and stored alongside the EntityGraph memories for that clip.

\subsection{EntityGraph: Entity-Centric Episodic and Semantic Memory}
\label{sec:entitygraph}
Many questions about long videos require reasoning about specific characters, such as tracking their actions, attributes, or relationships across scenes.
This motivates an identity-centric memory that organizes observations around recognized characters.
Following M3-Agent~\citep{long2026seeing}, we construct the EntityGraph $\mathcal{G}_\text{Ent} = (\mathcal{N}, \mathcal{E}_\text{Ent})$ incrementally as clips are processed.
The graph maintains two types of identity anchors: image nodes that store face embeddings, and voice nodes that store speaker embeddings.
For each clip $c_t$, a face module detects and clusters faces, and matches each cluster to existing image nodes by embedding similarity.
An audio module extracts speaker embeddings from ASR segments with speaker-turn boundaries, and matches them to existing voice nodes.
Around these identity anchors, two types of textual memories are accumulated.
Episodic memory nodes store clip-level observations, while semantic memory nodes store high-level knowledge, such as character identities, attributes, relationships, and general facts.
For each clip, a VLM generates episodic and semantic memories from the clip video, character identity tags, and diarized dialogue transcripts.
Each text memory node is connected to the identity anchors it mentions, so that retrieval can follow these links to gather all observations about a given character.
Cross-modal equivalences stated in semantic memories further unify image and voice tags into shared character identifiers, enabling identity-based retrieval even when a character is recognized through different modalities across clips.

\subsection{EventGraph: Event-Centric Narrative Memory}
\label{sec:eventgraph}

The EntityGraph organizes memories around individual characters, but many questions require reasoning about the relationships between events. For example, understanding why a character acts in a certain way may require tracing a causal chain across multiple scenes. Research in event cognition shows that humans naturally segment experience into discrete event units linked by temporal and causal dependencies~\citep{zacks2007event,hahamy2023human}. 
To capture this narrative structure, we introduce the \emph{EventGraph} $\mathcal{G}_E = (\mathcal{S}, \mathcal{R})$, a directed graph whose nodes are event-level segments and whose edges encode temporal and causal relations.

\noindent{\textbf{Event Segments.}}
Each segment node stores a short descriptive label (e.g., ``Kitchen Preparation''), a natural-language summary of the event, and the set of clip IDs it spans. The clip set also serves as a pointer into the lower memory layers. All entity-level memories and keyframes whose clip ID falls within the segment are accessible through it, enabling retrieval from coarse event context down to fine-grained details.

\noindent{\textbf{Temporal-Causal Edges.}}
Each directed edge connects two segments and carries a relation type (temporal or causal), a natural-language descriptor, evidence clip IDs, and a short rationale. Temporal edges capture ordering, while causal edges capture dependencies (e.g., ``enables'' or ``prevents''). Edges are not restricted to adjacent segments. Long-range links are allowed whenever supported by evidence, which is important for capturing event-level dependencies across distant clips.

\noindent{\textbf{Incremental Construction.}}
The EventGraph is constructed online as new clips arrive, without access to future content. After every consecutive group of clips, an LLM receives the most recently added segments and edges together with the newly extracted clip-level memories, and produces a structured graph update. The update supports three operations. \emph{Create} introduces a new segment for an emergent event. \emph{Extend} appends new clip IDs to an existing segment whose event is still ongoing and updates its summary. \emph{Connect} adds, modifies, or removes edges as new dependencies become apparent.
After each update is applied, a \emph{subset absorption} step merges any segment whose clip set is a strict subset of another into the larger segment and removes its edges. This prevents fragmentation and keeps the graph compact as the video progresses.
Passing only the most recent update rather than the full graph bounds the context length seen by the LLM while providing sufficient local structure for coherent continuation. The incremental design also allows the graph to evolve with the narrative, as early segments can be refined, merged, or re-connected when more context becomes available.

\subsection{Agentic Reasoning over Hierarchical Memory}
\label{sec:control}

When understanding long videos, humans do not scan their memory linearly. They first localize the relevant event, then retrieve finer details within it, and finally synthesize evidence from multiple sources. Our agentic reasoner follows the same strategy: it decomposes each question into focused subtasks and retrieves evidence from the appropriate memory level through multi-round interaction.
Algorithm~\ref{alg:agentic-reasoning} summarizes the full procedure, which we detail below.

\noindent{\textbf{Subtask Decomposition and Context Management.}}
Answering a question about a long video often requires collecting evidence across different time spans or modalities. A single retrieval query may miss evidence that is spread across distant scenes or encoded in different modalities. The reasoner therefore decomposes the question into several focused subtasks, each targeting a specific aspect of the required evidence.
To manage the reasoning context across multiple retrieval rounds, we maintain a structured task ledger $\mathcal{L}$ that records the dependencies, status, attempts, retrieved evidence, and provisional answer for every subtask.
A scheduler selects one unresolved subtask at each round, and the selected subtask becomes the current focus injected into the prompt. 
Because the ledger makes the reasoning state explicit and persistent, useful evidence is not lost even when the dialogue history grows beyond the model's effective attention span. The final synthesis also reads from the ledger rather than from the full conversation, so the answer is grounded in compact, task-aligned evidence regardless of how many rounds were needed.

\noindent{\textbf{Multilevel Retrieval with Harness Engineering.}}
At each round, the reasoner selects one of two actions: issuing a retrieval query to gather additional evidence, or producing a final answer when the accumulated evidence is deemed sufficient. Retrieval queries are routed to the three-layer memory hierarchy through four modes that operate at progressively finer granularity. An \texttt{EVENT} query searches the EventGraph and returns the best matching event segment together with its one-hop temporal-causal neighbors, providing coarse narrative context. A \texttt{VIDEO} query drills down within the current focus event and retrieves EntityGraph memories from the corresponding clips, surfacing entity-level details such as dialogue, actions, and character attributes. A \texttt{NEIGHBOR} query shifts the focus to a related event segment by following EventGraph edges, enabling multi-hop evidence gathering across causally or temporally linked events. A \texttt{KEYFRAME} query loads representative frames from the focused clips when the question requires visual evidence such as appearance, spatial layout, or on-screen text. The agent iterates through these modes until the evidence is sufficient.

To ensure that this multi-round retrieval loop executes reliably, we adopt the \emph{harness engineering} paradigm, which wraps the model with infrastructure that informs, constrains, verifies, and corrects its behavior throughout execution. We instantiate this paradigm through four complementary mechanisms, and show a case in Table~\ref{tab:harness_case_rick}.

\emph{Inform.}
Each retrieval information is transformed into a structured text summary before being appended to the dialogue. The formatter extracts the matched event segment, its temporal-causal neighbors, and any entity-level details, while filtering out metadata irrelevant to reasoning. In addition, a snapshot of the task ledger is injected alongside each retrieval result, presenting the current status of all subtasks, accumulated evidence, and the currently focused subtask. The model thus receives both the full dialogue history and a compact, up-to-date view of the reasoning state at every round.
This dual view allows the model to revisit raw evidence from any prior round while staying oriented toward the current subtask, preventing attention drift as the context grows.

\emph{Constrain.}
The harness enforces deterministic guardrails on every model output. An output format validator rejects malformed responses that lack a valid action structure or contain an illegal answer option, triggering an immediate retry. A query deduplication module computes word-level Jaccard similarity against the search history and blocks near-duplicate queries, forcing the model to diversify its retrieval strategy rather than repeating ineffective searches. A resource budget caps both wall-clock time and token consumption per question, and triggers early termination when consecutive retrieval rounds return no useful results. 

\emph{Verify.}
After each retrieval round, a lightweight sufficiency signal estimates whether the accumulated evidence is adequate. Each payload is reduced to a keyword fingerprint, and the fraction of previously unseen keywords measures how much new information the round contributed. When this fraction stays below a fixed threshold for consecutive rounds, the signal declares information saturation and advises the model to stop searching. This prevents the model from wasting retrieval rounds on diminishing returns while also providing the Correct mechanism with a trigger to intervene when progress stalls.

\emph{Correct.}
When a failure is detected, the harness intervenes at the appropriate level. Invalid outputs are fed back with the specific error for in-context retry. When the sufficiency signal fires, a grace round injects strategy suggestions before forcing an answer. When a subtask makes no progress after repeated attempts, the planner performs incremental repair, rewriting the stuck subtask or abandoning unproductive branches while preserving already-resolved ones. This graduated response matches each failure to the minimal intervention needed.

\begin{algorithm*}[tb]
   \caption{Agentic Reasoning over Hierarchical Memory}
   \label{alg:agentic-reasoning}
\begin{algorithmic}
   \STATE {\bfseries Input:} question $q$, answer choices $A$, hierarchical memory $\mathcal{M}$, round budget $R$
   \STATE {\bfseries Output:} final answer $y$
   \STATE Prepare memory state $\mathcal{M}_q$ for the current video.
   \STATE Route relevant skills and initialize task ledger $\mathcal{L}$ from $q$.
   \STATE Initialize progress ledger $\Pi$ and focus event $s \leftarrow \varnothing$.
   \FOR{$r = 1$ {\bfseries to} $R$}
      \STATE Evaluate the progress signal of $\Pi$ over $\mathcal{L}$.
      \IF{the signal calls for replanning}
         \STATE Repair $\mathcal{L}$: rewrite, add, or abandon unresolved subtasks.
      \ELSIF{all subtasks are resolved, none is schedulable, or the budget is exhausted}
         \STATE \textbf{break}
      \ENDIF
      \STATE Schedule an unresolved subtask $u$ from $\mathcal{L}$ as the current focus.
      \STATE Generate action $a \in \{\texttt{Answer}, \texttt{Search}\}$ conditioned on $q$, $u$, and $\mathcal{L}$.
      \STATE Validate $a$ with the harness guardrails (format, deduplication), retrying on violation.
      \IF{$a = \texttt{Answer}$}
         \STATE \textbf{return} the answer parsed from $a$.
      \ENDIF
      \STATE Parse search query $z$ and retrieval mode $m$.
      \IF{$m = \texttt{EVENT}$}
         \STATE Retrieve event segment $s$ and its temporal-causal neighbors from $\mathcal{G}_E$.
      \ELSIF{$m = \texttt{VIDEO}$}
         \STATE Retrieve entity-level memories from clips covered by $s$.
      \ELSIF{$m = \texttt{NEIGHBOR}$}
         \STATE Move $s$ to a related event through edges in $\mathcal{G}_E$.
      \ELSIF{$m = \texttt{KEYFRAME}$}
         \STATE Retrieve keyframes from the perceptual buffer $\mathcal{P}$ within clips covered by $s$.
      \ENDIF
      \STATE Attach retrieved evidence to $u$ and update $\mathcal{L}$.
      \STATE Assess information sufficiency and inject a strategy hint when saturated.
      \STATE Record the round outcome into $\Pi$.
      \IF{$s$ stays locked on with no new evidence}
         \STATE Force the next retrieval mode (\texttt{VIDEO}, then \texttt{NEIGHBOR}).
      \ENDIF
   \ENDFOR
   \STATE Synthesize $y$ from the final task ledger $\mathcal{L}$.
   \STATE Record the trajectory and outcome for self-evolution.
\end{algorithmic}
\end{algorithm*}

\emph{Self-Evolving.}
Beyond controlling a single reasoning trajectory, the harness also accumulates experience across questions. After each question is answered, a learning capture module automatically extracts structured observations from the trajectory and evaluation outcome, recording which retrieval strategies succeeded or failed for each question type. Once observations of the same pattern accumulate beyond a threshold within a batch, they are promoted into a reusable skill equipped with specific instructions for decomposition, mode selection, or answer calibration. At inference time, a skill router matches incoming questions to existing skills by embedding similarity and injects the matched instructions into the agent's prompt. Skills that degrade downstream accuracy are automatically disabled. Because this loop operates entirely on the reasoning policy, the underlying memory graphs remain unchanged and the system improves without retraining any model.

\section{Experiments}
\subsection{Experimental Details}

\paragraph{Benchmarks.}
We evaluate the long video understanding on challenging benchmarks.
\textbf{Video-MME-long}~\citep{fu2025video} consists of 900 questions based on 300 videos ranging from 30–60 min.
\textbf{M3-Bench} consists of robot and web subsets, drawn from robot-perspective real-life scenario videos and YouTube videos respectively. The robot subset contains 1,276 questions over 100 videos averaging 35 minutes in length. The web subset contains 3,214 questions over 900 videos averaging 27 minutes in length. 
M3-Bench uses open-ended questions. We adopt Gemini-2.5-Flash as an LLM judge. The other two datasets use multiple-choice questions. We adopt rule-based evaluation for them. In addition, to evaluate omni-modal video understanding, we do not use external subtitle information.

\paragraph{Baselines.}
We compare our method with the following competitive online long video understanding methods:
 \begin{itemize} [leftmargin=*,itemsep=-1pt]
\setlength{\itemsep}{0pt}
\setlength{\parskip}{0pt}
    \item MovieChat~\citep{song2024moviechat} extracts short-term and long-term visual features using a sliding window, then projects the visual representations to an LLM for video understanding.
     \item MA-LMM~\citep{he2024malmm} employs a long-term visual memory bank that extracts frame features online and compresses visual tokens through temporal modeling to maintain memory capacity.
    \item Flash-VStream~\citep{zhang2024flashvstream} introduces a Flash Memory module for efficient long video understanding. The module pairs a low-capacity context memory that aggregates temporal information and information density with a augmentation memory that retrieves spatial details conditioned on the density distribution.
    \item M3-Agent~\citep{long2026seeing} introduces a multimodal agent framework equipped with long-term memory. The framework organizes episodic and semantic memories in an entity-centric manner, builds and updates them from real-time visual and auditory inputs, and enables autonomous multi-turn reasoning and memory retrieval to complete tasks.
    \item Gemini-Agent \& Gemini-GPT4o-Hybrid adopt the same memory-control agent framework as M3-Agent. Gemini-Agent uses Gemini-1.5-pro for both memory construction and control. Gemini-GPT4o-Hybrid uses GPT4o as the control model.
\end{itemize}

\paragraph{Implementation details.}
Videos are segmented into 30-second clips. For each clip, the Perceptual Buffer retains 3 keyframes selected by the hybrid shot-boundary and quality scoring procedure described in Section~\ref{sec:perceptual}. Face detection and embedding use InsightFace. 
Speaker diarization and ASR are performed by Qwen3-Omni, and speaker embeddings are extracted with ERes2NetV2. EntityGraph episodic and semantic memories are then generated by M3-Agent-Memorization, a fine-tuned Qwen2.5-Omni model that receives the clip video, detected face tags, and diarized dialogue transcripts. The EventGraph is constructed incrementally with Gemini-3-Flash-Preview, updating every 10 clips using the sliding-window patch protocol described in Section~\ref{sec:eventgraph}. 
At question time, the reasoning agent uses Gemini-3-Flash-preview with temperature 0.8. The task ledger allows up to 5 subtasks, each with at most 3 retrieval attempts, and the round budget is set to 10. Each \texttt{KEYFRAME} retrieval returns at most 5 images. For self-evolution, trajectories are accumulated within batches of 64 questions. 
A pattern is promoted to a reusable skill when it recurs at least 8 times, and skills are routed to new questions when the embedding similarity exceeds 0.45. Apart from the fine-tuned memorization model, the entire reasoning framework is training-free.

\subsection{Main Results}

\begin{table*}[t]
\centering
\resizebox{\textwidth}{!}{
\begin{tabular}{lcccccccccccccc}
\toprule
\multirow{2}{*}{Method}
  & \multicolumn{6}{c}{M3-Bench-robot}
  & \multicolumn{6}{c}{M3-Bench-web}
  & \multirow{2}{*}{\makecell{Video-\\MME-L}} \\
\cmidrule(lr){2-7}\cmidrule(lr){8-13}
  & ME & MH & CM & PU & GK & \textbf{All}
  & ME & MH & CM & PU & GK & \textbf{All} & & \\
\midrule
\multicolumn{14}{c}{\textit{Online Video Understanding Methods}} \\
\midrule
MovieChat            & 13.3 & 9.8  & 12.2 & 15.7 & 7.0  & 11.2 & 12.2 & 6.6  & 12.5 & 17.4 & 11.1 & 12.6 & 19.4 \\
MA-LMM               & 25.6 & 23.4 & 22.7 & 39.1 & 14.4 & 24.4 & 26.8 & 10.5 & 22.4 & 39.3 & 15.8 & 24.3 & 17.3  \\
Flash-VStream        & 21.6 & 19.4 & 19.3 & 24.3 & 14.1 & 19.4 & 24.5 & 10.3 & 24.6 & 32.5 & 20.2 & 23.6 & 25.0  \\
\midrule
\multicolumn{14}{c}{\textit{Agent Method}} \\
\midrule
Gemini-Agent         & 15.8 & 17.1 & 15.3 & 20.0 & 15.5 & 16.9 & 29.3 & 20.9 & 33.8 & 34.6 & 45.0 & 34.1 & 55.1  \\
Gemini-GPT-Hybrid  & 21.3 & 25.5 & 22.7 & 28.8 & 23.1 & 24.0 & 35.9 & 26.2 & 37.6 & 43.8 & 52.2 & 41.2 & 56.5  \\
M3-Agent    & 32.8 & 29.4 & 31.2 & 43.3 & 19.1 & 30.7 & 45.9 & 28.4 & 44.3 & 59.3 & 53.9 & 48.9 & 61.8  \\
M3-Agent-Gemini25 & 24.6 & 40.0 & 23.3 & 32.1 & 19.9 & 24.6 & 37.0 & 26.5 & 39.2 & 47.9 & 53.1 & 43.0 & 55.5 \\
M3-Agent-Gemini3 & 38.5 & 37.7 & 35.1 & 48.5 & 27.2 & 37.3 & 40.3 & 29.3 & 37.7 & 50.7 & 55.3 & 45.6 & 70.7 \\
\textsc{Homer} (Gemini3) & \textbf{43.7} & \textbf{42.4} & \textbf{43.3} & \textbf{60.4} & \textbf{28.1} & \textbf{42.8} & \textbf{55.3} & \textbf{42.0} & \textbf{50.7} & \textbf{65.0} & \textbf{59.1} & \textbf{56.4} & \textbf{75.1} \\
\bottomrule
\end{tabular}
}
\vspace{-10pt}
\caption{Main results on M3-Bench and Video-MME-Long. We compare \textsc{Homer} with online video understanding models and agent-based baselines. ME, MH, CM, PU, GK denote multi-evidence reasoning, multi-hop reasoning, cross-modal reasoning, person understanding, and general knowledge extraction question types, respectively. All numbers are accuracy in \%, and \textbf{All} reports the overall accuracy on the full split.}

\label{tab:main}
\end{table*}

Table~\ref{tab:main} compares \textsc{Homer} with online video understanding methods and agent-based methods on M3-Bench and Video-MME-Long. 
The five M3-Bench categories are abbreviated as ME for multi-evidence reasoning, MH for multi-hop reasoning, CM for cross-modal reasoning, PU for person understanding, and GK for general knowledge extraction.
\textsc{Homer} achieves the best result on every benchmark, reaching $42.8\%$ on M3-Bench-robot, $56.4\%$ on M3-Bench-web, and $75.1\%$ on Video-MME-Long, which corresponds to $+5.5$, $+10.8$, and $+4.4$ points over the previous best agent method M3-Agent-Gemini3. 
The lift is largest on M3-Bench-web, whose videos span broader topics and longer narratives, suggesting that the hierarchical memory and the harness pay off more as the search space grows and questions demand evidence from more distant moments.
MovieChat, MA-LMM, and Flash-VStream represent online methods that maintain compact visual or KV-level memories, and their accuracy on M3-Bench-robot stays between $11.2\%$ and $24.4\%$, leaving a gap of at least $18$ points to \textsc{Homer}. 
The gap is most pronounced on person understanding and cross-modal reasoning, since compressed visual features cannot recover entity identities or align visual evidence with dialogue once they have been squeezed out of the representation. 
For long-form videos with complex narrative structure, semantic-level abstractions over the stream rather than raw feature compression are what enable fine-grained evidence retrieval.

Compared with M3-Agent-Gemini3, which shares the same memory construction pipeline and Gemini-3-Flash backbone but lacks the EventGraph and the harness, \textsc{Homer}'s per-category gains on M3-Bench-robot concentrate on person understanding with $+11.9$ and cross-modal reasoning with $+8.2$. 
The former reflects the EntityGraph, which keeps a persistent identity for each character and accumulates appearance, dialogue, and action evidence across scenes. 
The latter reflects the hierarchical memory storing visual frames alongside diarized speech and ASR transcripts, so that the agent retrieves from either modality without a separate alignment step. 
By contrast, general knowledge extraction lifts by only $+0.9$, indicating that \textsc{Homer} contributes structure for organizing what the video itself provides rather than additional knowledge on top of the language model.

\section{Analysis}
\subsection{How Much Does Each Memory Layer Contribute?}

We isolate each memory layer by incrementally building \textsc{Homer} from the EntityGraph upward, evaluating on a random subset of $600$ questions from M3-Bench-robot (Table~\ref{tab:memory_ablation}). Starting from the EntityGraph baseline ($37.8$ overall), we examine how each additional layer introduced by \textsc{Homer} reshapes the performance profile.

\begin{table}[t]
\centering
\small
\resizebox{0.5\textwidth}{!}{%
\begin{tabular}{l cccccc}
\toprule
Memory & ME & MH & CM & PU & GK & \textbf{All} \\
\midrule
EntityGraph only                              & 39.6 & 41.5 & 36.2 & 49.0 & 22.2 & 37.8 \\
+ Event Nodes                           & 41.8 & 51.2 & 39.8 & 56.6 & 27.0 & 41.2 \\
+ Event Graph                           & 46.1 & 43.9 & 43.3 & 59.9 & 28.6 & 44.5 \\
+ Perceptual Buffer    & 45.6 & 43.9 & 44.1 & 57.3 & 27.8 & 44.0 \\
\bottomrule
\end{tabular}
}
\vspace{-10pt}
\caption{Memory ablation on M3-Bench-robot.}
\label{tab:memory_ablation}
\end{table}

Adding event nodes on top of the EntityGraph produces the largest single gain, lifting multi-hop reasoning from $41.5$ to $51.2$. Event segments give the agent explicit temporal anchors for chaining evidence across scenes. Promoting these nodes into a full Event Graph by adding temporal and causal edges then shifts the gain pattern: multi-evidence and cross-modal reasoning improve, while multi-hop drops by $7.3$ points. The edges introduce graph-walk retrieval, which expands the agent's reachable context per round. This benefits questions with a clear causal chain but introduces a locality bias, where the agent preferentially traverses nearby edges rather than issuing a fresh semantic query to reach distant target events.
Stacking the Perceptual Buffer on top of the Event Graph produces a targeted effect. Cross-modal reasoning rises by $0.8$ points because some questions demand raw visual evidence, while other categories drift by less than a point. The textual hierarchy of entity and event memory already covers most of the reasoning surface, and the buffer serves as a safety net for visually grounded questions rather than a default retrieval channel.
In summary, the EventGraph contributes the dominant structural gain introduced by \textsc{Homer}, with causal edges trading exploratory breadth for precision on causally structured questions. The Perceptual Buffer complements this with selective visual grounding.

\subsection{Does \textsc{Homer} Generalize across Backbones?}

We replace the reasoning agent with three alternative LLMs and compare each against the corresponding M3-Agent baseline on M3-Bench-robot. Two of them, Deepseek-v4-Flash and Qwen3-Coder-480B-A35B, are text-only and never see a frame from the Perceptual Buffer. 
As shown in Table~\ref{tab:harness_models}, \textsc{Homer} lifts every backbone, taking Gemini-2.5-Flash from $24.6$ to $30.6$, Deepseek-v4-Flash from $25.1$ to $33.9$, and Qwen3-Coder from $19.4$ to $23.8$. The per-category profile is consistent across all three: person understanding receives the largest gain while general knowledge extraction barely moves. 
Because this profile persists on text-only backbones, the gain comes from the hierarchical memory itself rather than the LLM's visual capability, making \textsc{Homer}'s contribution structural and model-agnostic.

\begin{table}[t]
\centering
\label{tab:harness_models}
\small
\resizebox{0.5\textwidth}{!}{%
\begin{tabular}{l cccccc}
\toprule
Backbone & ME & MH & CM & PU & GK & \textbf{All} \\
\midrule
Gemini-2.5-Flash             & 24.6 & 40.0 & 23.3 & 32.1 & 19.9 & 24.6 \\
+ \textsc{Homer}     & 31.0 & 35.3 & 30.9 & 44.2 & 21.1 & \textbf{30.6} \\
\midrule
Deepseek-v4-Flash             & 25.7   & 24.7   & 23.5   & 30.3   & 22.0   & 25.1 \\
 + \textsc{Homer}     & 34.4   & 31.8   & 33.8   & 47.5   & 22.3   & \textbf{33.9} \\
\midrule
Qwen3-Coder-480B             & 18.8   & 16.5   & 19.5   & 22.6   & 19.6   & 19.4 \\
+ \textsc{Homer}     & 24.1   & 21.2   & 22.7   & 34.5   & 17.1   & \textbf{23.8} \\
\bottomrule
\end{tabular}
}
\vspace{-10pt}
\caption{Effect of \textsc{Homer} on M3-Bench-robot under three different reasoning backbones.}
\end{table}

The relative ordering of backbones also shifts under \textsc{Homer}: Gemini-2.5-Flash and Deepseek-v4-Flash are tied as M3-Agent baselines ($24.6$ vs.\ $25.1$), yet within \textsc{Homer} Deepseek pulls ahead to $33.9$ while Gemini-2.5 reaches $30.6$, indicating that the bottleneck is not the LLM's raw video understanding but its ability to drive a structured retrieval-and-reasoning loop. Without any video-specific fine-tuning, the mid-scale open-source Qwen3-Coder reaches $23.8$ overall and $34.5$ on person understanding inside \textsc{Homer}, matching the closed-source multimodal Gemini-2.5-Flash.

\begin{figure*}[t]
\centering
\includegraphics[width=\textwidth]{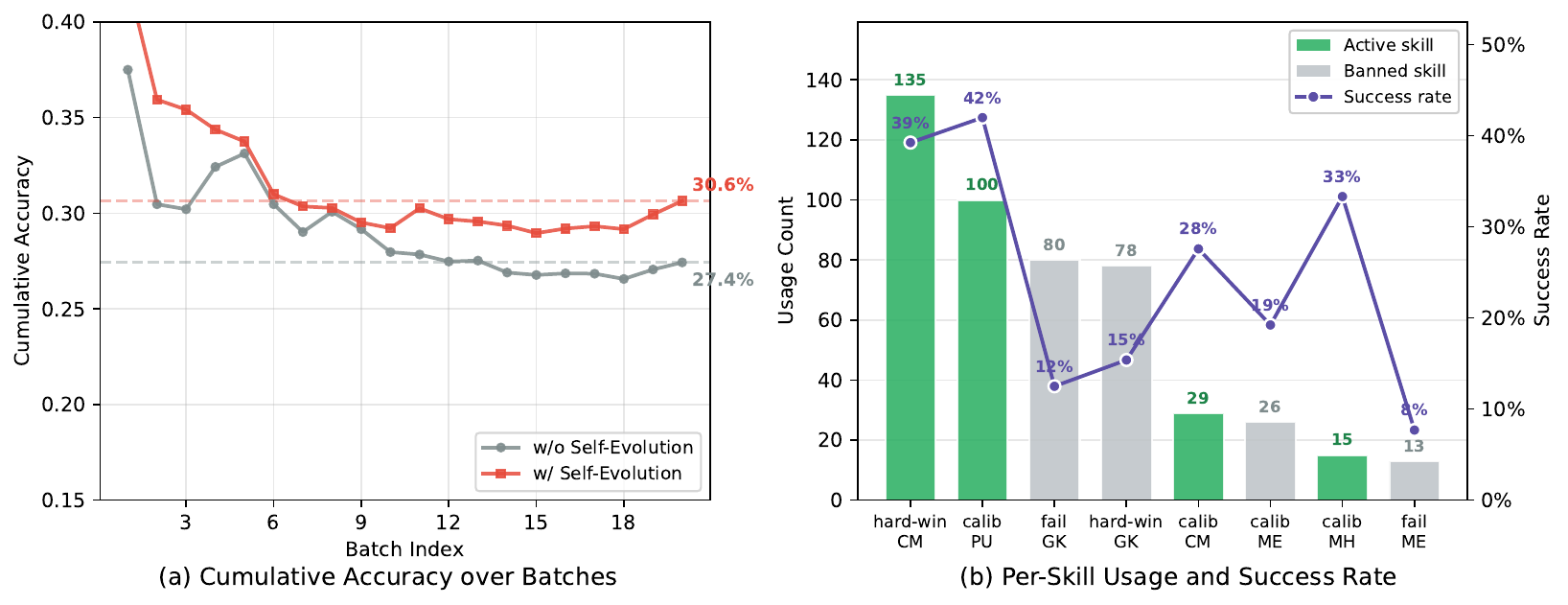}
\vspace{-20pt} 
\caption{Self-evolution behaviour of \textsc{Homer} on M3-Bench-robot, processed in batches of $64$ questions. (a) Cumulative accuracy with and without self-evolution; (b) per-skill usage and downstream success rate, where green skills remain active and grey ones are automatically banned once their success rate falls below the running baseline.}
\label{fig:self_evolution_curve}
\end{figure*}

\subsection{How Does the Agent Evolve over Time?}

To examine whether \textsc{Homer}'s self-evolution mechanism accumulates useful retrieval skills, we compare two \textsc{Homer} variants on the full M3-Bench-robot with the same Gemini-2.5-Flash backbone and identical memory: self-evolution disabled and batch-level self-evolution enabled. 
After each batch of $64$ questions, the harness consolidates failure- and success-mode signals into a question-type-aware skill library available to subsequent batches. 
Overall accuracy rises from $27.4\%$ to $30.6\%$ ($+11.7\%$ relative), and Figure~\ref{fig:self_evolution_curve}(a) shows that the evolving run pulls clear of the static baseline as more batches are processed, with the gap widening rather than closing in the second half, indicating that the gain arises from the accumulated library rather than from easier later questions.

The skill library captures reusable patterns rather than memorized question--answer pairs. Raw trajectories are consolidated into $14$ reusable skills along two axes, question type and signal type. 
The signal-type axis comprises calibration skills that sharpen confidence on systematically over-confident answers, hard-win skills that capture successful retrieval patterns, search-fail skills that record alternatives when retrieval saturates, and an error skill for repeated synthesis mistakes. 
The harness tracks downstream success rates and retires ineffective skills automatically. Figure~\ref{fig:self_evolution_curve}(b) shows two dominant active skills: a person-understanding calibration skill triggered $100$ times at a $42.0\%$ success rate, and a cross-modal hard-win skill triggered $135$ times at $39.3\%$. 
These question types coincide with the largest per-category gains in Table~\ref{tab:main}, which traces the macro improvement to specific reusable skills. By contrast, every skill mined for general knowledge extraction falls below the running baseline and is automatically banned, consistent with the negligible GK gain in Table~\ref{tab:main} and confirming that retrieval skills cannot substitute for general knowledge. 
The library therefore remains compact and self-correcting even after thousands of trajectories, allowing the agent to keep improving without retraining the underlying LLM.

\subsection{Where Does the System Fail?}

To understand the remaining errors, we attribute each of the $2{,}131$ incorrect trajectories on M3-Bench to the first stage where the trajectory went wrong. 
The pipeline has four stages: memory construction, retrieval, event localization, and answer synthesis. An LLM judge inspects the question, ground-truth answer, model prediction, search queries, per-round retrieval excerpts, and harness signals, then assigns one of four labels: \emph{Memory (absent)}, \emph{Memory (coarse)}, \emph{Event localization}, and \emph{Answer synthesis}.

\begin{figure}[t]
\centering
\includegraphics[width=0.96\columnwidth]{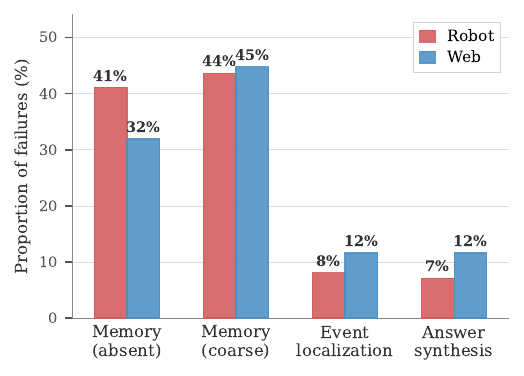}
\vspace{-15pt}
\caption{Failure attribution by LLM-as-judge on M3-Bench-robot and M3-Bench-web.}
\label{fig:error_attribution}
\end{figure}

As shown in Figure~\ref{fig:error_attribution}, the two memory-related labels together account for over $80\%$ of errors on robot and $77\%$ on web. 
Roughly half are cases where the agent searched the right region of memory but the stored representation lacked the decisive fine-grained detail, and the rest are cases where the relevant fact was never written into memory in the first place. Both sub-types point to the same conclusion: the agent's search behavior is largely correct, and the real bottleneck lies upstream in what gets written into memory. The remaining two categories account for a much smaller share of errors.
Event localization ($8\%$ / $12\%$) reflects the agent persistently focusing on the wrong event segment, while answer synthesis errors ($7\%$ / $12\%$) arise when sufficient evidence is retrieved but the final response confuses entities or adds unsupported details. Future gains should therefore prioritize denser and more accurate episodic summaries during 
memory construction, with secondary improvements from constraining the final synthesis to stay faithful to retrieved evidence.

\section{Conclusion}
We presented \textsc{Homer}, a hierarchical online memory framework for long-form video understanding that pairs a perceptual--entity--event memory with an agentic reasoner supervised by a verify-and-correct harness. Across M3-Bench and Video-MME-Long, \textsc{Homer} consistently surpasses prior agent methods, with the gain transferring to three reasoning backbones and letting an open-source agent match a closed-source multimodal model. Our failure analysis shows that the dominant remaining bottleneck lies in memory construction rather than downstream reasoning, suggesting that richer episodic representations offer the most direct path to further gains.

\bibliography{tacl2021}

@inproceedings{
shen2025vgent,
title={Vgent: Graph-based Retrieval-Reasoning-Augmented Generation For Long Video Understanding},
author={Xiaoqian Shen and Wenxuan Zhang and Jun Chen and Mohamed Elhoseiny},
booktitle={The Thirty-ninth Annual Conference on Neural Information Processing Systems},
year={2025},
url={https://openreview.net/forum?id=5xPvWat3IX}
}

@inproceedings{wang2025videotree,
  title={Videotree: Adaptive tree-based video representation for llm reasoning on long videos},
  author={Wang, Ziyang and Yu, Shoubin and Stengel-Eskin, Elias and Yoon, Jaehong and Cheng, Feng and Bertasius, Gedas and Bansal, Mohit},
  booktitle={Proceedings of the Computer Vision and Pattern Recognition Conference},
  pages={3272--3283},
  year={2025}
}

@inproceedings{tang2025aks,
  title={Adaptive keyframe sampling for long video understanding},
  author={Tang, Xi and Qiu, Jihao and Xie, Lingxi and Tian, Yunjie and Jiao, Jianbin and Ye, Qixiang},
  booktitle={Proceedings of the Computer Vision and Pattern Recognition Conference},
  pages={29118--29128},
  year={2025}
}

@article{zhu2025focus,
  title={Focus: Efficient keyframe selection for long video understanding},
  author={Zhu, Zirui and Xu, Hailun and Luo, Yang and Liu, Yong and Sarkar, Kanchan and Yang, Zhenheng and You, Yang},
  journal={arXiv preprint arXiv:2510.27280},
  year={2025}
}

@article{zemskova2026focusgraph,
  title={FocusGraph: Graph-Structured Frame Selection for Embodied Long Video Question Answering},
  author={Zemskova, Tatiana and Andryushenko, Solomon and Obrubov, Ilya and Khoruzhaia, Viktoriia and Eroshenko, Ekaterina and Derevyanka, Ekaterina and Yudin, Dmitry},
  journal={arXiv preprint arXiv:2603.04349},
  year={2026}
}

@inproceedings{weng2024longvlm,
  title={Longvlm: Efficient long video understanding via large language models},
  author={Weng, Yuetian and Han, Mingfei and He, Haoyu and Chang, Xiaojun and Zhuang, Bohan},
  booktitle={European Conference on Computer Vision},
  pages={453--470},
  year={2024},
  organization={Springer}
}

@inproceedings{wang2024videoagent,
  title={Videoagent: Long-form video understanding with large language model as agent},
  author={Wang, Xiaohan and Zhang, Yuhui and Zohar, Orr and Yeung-Levy, Serena},
  booktitle={European Conference on Computer Vision},
  pages={58--76},
  year={2024},
  organization={Springer}
}

@inproceedings{ma2025drvideo,
  title={Drvideo: Document retrieval based long video understanding},
  author={Ma, Ziyu and Gou, Chenhui and Shi, Hengcan and Sun, Bin and Li, Shutao and Rezatofighi, Hamid and Cai, Jianfei},
  booktitle={Proceedings of the Computer Vision and Pattern Recognition Conference},
  pages={18936--18946},
  year={2025}
}

@inproceedings{
liu2026videomind,
title={VideoMind: A Chain-of-Lo{RA} Agent for Temporal-Grounded Video Reasoning},
author={Ye Liu and Kevin Qinghong Lin and Chang Wen Chen and Mike Zheng Shou},
booktitle={The Fourteenth International Conference on Learning Representations},
year={2026},
url={https://openreview.net/forum?id=57EwidOnSf}
}

@article{longvt2026,
  title={Longvt: Incentivizing" thinking with long videos" via native tool calling},
  author={Yang, Zuhao and Wang, Sudong and Zhang, Kaichen and Wu, Keming and Leng, Sicong and Zhang, Yifan and Li, Bo and Qin, Chengwei and Lu, Shijian and Li, Xingxuan and others},
  journal={arXiv preprint arXiv:2511.20785},
  year={2025}
}

@inproceedings{lin2025vlog,
  title={Vlog: Video-language models by generative retrieval of narration vocabulary},
  author={Lin, Kevin Qinghong and Shou, Mike Zheng},
  booktitle={Proceedings of the IEEE/CVF Conference on Computer Vision and Pattern Recognition},
  pages={3218--3228},
  year={2025}
}

@inproceedings{li2024llamavid,
  title={Llama-vid: An image is worth 2 tokens in large language models},
  author={Li, Yanwei and Wang, Chengyao and Jia, Jiaya},
  booktitle={European Conference on Computer Vision},
  pages={323--340},
  year={2024},
  organization={Springer}
}

@inproceedings{
shen2024longvu,
title={Long{VU}: Spatiotemporal Adaptive Compression for Long Video-Language Understanding},
author={Xiaoqian Shen and Yunyang Xiong and Changsheng Zhao and Lemeng Wu and Jun Chen and Chenchen Zhu and Zechun Liu and Fanyi Xiao and Balakrishnan Varadarajan and Florian Bordes and Zhuang Liu and Hu Xu and Hyunwoo J. Kim and Bilge Soran and Raghuraman Krishnamoorthi and Mohamed Elhoseiny and Vikas Chandra},
booktitle={Forty-second International Conference on Machine Learning},
year={2025},
url={https://openreview.net/forum?id=XzZC4gs1mf}
}

@article{chen2025streamingtom,
  title={Streamingtom: Streaming token compression for efficient video understanding},
  author={Chen, Xueyi and Tao, Keda and Shao, Kele and Wang, Huan},
  journal={arXiv preprint arXiv:2510.18269},
  year={2025}
}

@inproceedings{song2024moviechat,
  title={Moviechat: From dense token to sparse memory for long video understanding},
  author={Song, Enxin and Chai, Wenhao and Wang, Guanhong and Zhang, Yucheng and Zhou, Haoyang and Wu, Feiyang and Chi, Haozhe and Guo, Xun and Ye, Tian and Zhang, Yanting and others},
  booktitle={Proceedings of the IEEE/CVF Conference on Computer Vision and Pattern Recognition},
  pages={18221--18232},
  year={2024}
}

@inproceedings{he2024malmm,
  title={Ma-lmm: Memory-augmented large multimodal model for long-term video understanding},
  author={He, Bo and Li, Hengduo and Jang, Young Kyun and Jia, Menglin and Cao, Xuefei and Shah, Ashish and Shrivastava, Abhinav and Lim, Ser-Nam},
  booktitle={Proceedings of the IEEE/CVF conference on computer vision and pattern recognition},
  pages={13504--13514},
  year={2024}
}

@inproceedings{zhang2024flashvstream,
  title={Flash-vstream: Efficient real-time understanding for long video streams},
  author={Zhang, Haoji and Wang, Yiqin and Tang, Yansong and Liu, Yong and Feng, Jiashi and Jin, Xiaojie},
  booktitle={Proceedings of the IEEE/CVF international conference on computer vision},
  pages={21059--21069},
  year={2025}
}

@inproceedings{wang2025videollamb,
  title={Videollamb: Long streaming video understanding with recurrent memory bridges},
  author={Wang, Yuxuan and Song, Yiqi and Xie, Cihang and Liu, Yang and Zheng, Zilong},
  booktitle={Proceedings of the IEEE/CVF International Conference on Computer Vision},
  pages={24170--24181},
  year={2025}
}

@inproceedings{
streamforest2025,
title={StreamForest: Efficient Online Video Understanding with Persistent Event Memory},
author={Xiangyu Zeng and Kefan Qiu and Qingyu Zhang and Xinhao Li and Jing Wang and Jiaxin Li and Ziang Yan and Kun Tian and Meng Tian and Xinhai Zhao and Yi Wang and Limin Wang},
booktitle={The Thirty-ninth Annual Conference on Neural Information Processing Systems},
year={2025},
url={https://openreview.net/forum?id=9loSPaBwGO}
}

@article{xie2026fluxmem,
  title={FluxMem: Adaptive Hierarchical Memory for Streaming Video Understanding},
  author={Xie, Yiweng and He, Bo and Wang, Junke and Zheng, Xiangyu and Ye, Ziyi and Wu, Zuxuan},
  journal={arXiv preprint arXiv:2603.02096},
  year={2026}
}

@article{yang2025streammem,
  title={Streammem: Query-agnostic kv cache memory for streaming video understanding},
  author={Yang, Yanlai and Zhao, Zhuokai and Shukla, Satya Narayan and Singh, Aashu and Mishra, Shlok Kumar and Zhang, Lizhu and Ren, Mengye},
  journal={arXiv preprint arXiv:2508.15717},
  year={2025}
}

@article{tulving1972episodic,
  title={Episodic and semantic memory},
  author={Tulving, Endel and others},
  journal={Organization of memory},
  volume={1},
  number={381-403},
  pages={1},
  year={1972},
  publisher={New York}
}

@article{zacks2007event,
  title={Event perception: a mind-brain perspective.},
  author={Zacks, Jeffrey M and Speer, Nicole K and Swallow, Khena M and Braver, Todd S and Reynolds, Jeremy R},
  journal={Psychological bulletin},
  volume={133},
  number={2},
  pages={273},
  year={2007},
  publisher={American Psychological Association}
}

@article{hahamy2023human,
  title={The human brain reactivates context-specific past information at event boundaries of naturalistic experiences},
  author={Hahamy, Avital and Dubossarsky, Haim and Behrens, Timothy EJ},
  journal={Nature neuroscience},
  volume={26},
  number={6},
  pages={1080--1089},
  year={2023},
  publisher={Nature Publishing Group US New York}
}

@article{schacter1990perceptual,
  title={Perceptual representation systems and implicit memory: Toward a resolution of the multiple memory systems debate.},
  author={Schacter, Daniel L},
  journal={Annals of the New York Academy of Sciences},
  year={1990},
  publisher={New York Academy of Sciences}
}

@inproceedings{
long2026seeing,
title={Seeing, Listening, Remembering, and Reasoning: A Multimodal Agent with Long-Term Memory},
author={Lin Long and Yichen He and Wentao Ye and Yiyuan Pan and Yuan Lin and Hang Li and Junbo Zhao and Wei Li},
booktitle={The Fourteenth International Conference on Learning Representations},
year={2026},
url={https://openreview.net/forum?id=PMz29A7Muq}
}

@article{shen2026videovla,
  title={Videovla: Video generators can be generalizable robot manipulators},
  author={Shen, Yichao and Wei, Fangyun and Du, Zhiying and Liang, Yaobo and Lu, Yan and Yang, Jiaolong and Zheng, Nanning and Guo, Baining},
  journal={Advances in neural information processing systems},
  volume={38},
  pages={95597--95621},
  year={2026}
}

@article{pai2025mimic,
  title={mimic-video: Video-action models for generalizable robot control beyond vlas},
  author={Pai, Jonas and Achenbach, Liam and Montesinos, Victoriano and Forrai, Benedek and Mees, Oier and Nava, Elvis},
  journal={arXiv preprint arXiv:2512.15692},
  year={2025}
}

@inproceedings{chen2025gui,
  title={Gui-world: A video benchmark and dataset for multimodal gui-oriented understanding},
  author={Chen, Dongping and Huang, Yue and Wu, Siyuan and Tang, Jingyu and Zhou, Huichi and Zhang, Qihui and He, Zhigang and Bai, Yilin and Gao, Chujie and Chen, Liuyi and others},
  booktitle={International Conference on Learning Representations},
  volume={2025},
  pages={22812--22864},
  year={2025}
}

@inproceedings{ren2025videoworld,
  title={Videoworld: Exploring knowledge learning from unlabeled videos},
  author={Ren, Zhongwei and Wei, Yunchao and Guo, Xun and Zhao, Yao and Kang, Bingyi and Feng, Jiashi and Jin, Xiaojie},
  booktitle={Proceedings of the Computer Vision and Pattern Recognition Conference},
  pages={29029--29039},
  year={2025}
}

@article{zou2024seconds,
  title={From seconds to hours: Reviewing multimodal large language models on comprehensive long video understanding},
  author={Zou, Heqing and Luo, Tianze and Xie, Guiyang and Lv, Fengmao and Wang, Guangcong and Chen, Junyang and Wang, Zhuochen and Zhang, Hansheng and Zhang, Huaijian and others},
  journal={arXiv preprint arXiv:2409.18938},
  year={2024}
}

@article{bai2025qwen3,
  title={Qwen3-vl technical report},
  author={Bai, Shuai and Cai, Yuxuan and Chen, Ruizhe and Chen, Keqin and Chen, Xionghui and Cheng, Zesen and Deng, Lianghao and Ding, Wei and Gao, Chang and Ge, Chunjiang and others},
  journal={arXiv preprint arXiv:2511.21631},
  year={2025}
}

@article{dou2026cl,
  title={CL-bench Life: Can Language Models Learn from Real-Life Context?},
  author={Dou, Shihan and Shen, Yujiong and Huang, Chenhao and Ye, Junjie and Chen, Jiayi and Wang, Junzhe and He, Qianyu and Liu, Shichun and Lv, Changze and Lin, Jiahang and others},
  journal={arXiv preprint arXiv:2604.27043},
  year={2026}
}

@article{tang2025revisiting,
  title={Revisiting Long-context Modeling from Context Denoising Perspective},
  author={Tang, Zecheng and Ji, Baibei and Li, Juntao and Wu, Lijun and Gui, Haijia and Zhang, Min},
  journal={arXiv preprint arXiv:2510.05862},
  year={2025}
}

@inproceedings{fu2025video,
  title={Video-mme: The first-ever comprehensive evaluation benchmark of multi-modal llms in video analysis},
  author={Fu, Chaoyou and Dai, Yuhan and Luo, Yongdong and Li, Lei and Ren, Shuhuai and Zhang, Renrui and Wang, Zihan and Zhou, Chenyu and Shen, Yunhang and Zhang, Mengdan and others},
  booktitle={Proceedings of the IEEE/CVF conference on computer vision and pattern recognition},
  pages={24108--24118},
  year={2025}
}

@article{yan2026pyravid,
  title={PyraVid: Hierarchical Multimodal Memory for Long-Horizon Video Reasoning},
  author={Yan, Sikuan and Dong, Sicheng and Wang, Haotong and Nie, Ercong and Liu, Yilun and Bi, Jinhe and Xu, Yingjie and Schwarzmann, Susanna and Trivisonno, Riccardo and Tresp, Volker and others},
  journal={arXiv preprint arXiv:2605.17065},
  year={2026}
}

@article{lian2026verbatim,
  title={From Verbatim to Gist: Distilling Pyramidal Multimodal Memory via Semantic Information Bottleneck for Long-Horizon Video Agents},
  author={Lian, Niu and Wang, Yuting and Yao, Hanshu and Wang, Jinpeng and Chen, Bin and Wang, Yaowei and Zhang, Min and Xia, Shu-Tao},
  journal={arXiv preprint arXiv:2603.01455},
  year={2026}
}
\bibliographystyle{acl_natbib}










\appendix

\begin{table*}[t]
\centering
\small
\label{tab:harness_case_rick}
\begin{tabular}{p{0.02\textwidth} p{0.28\textwidth} p{0.52\textwidth} p{0.10\textwidth}}
\toprule
\# & Agent action & Harness intervention & Mechanism \\
\midrule
1 & \texttt{Search: Ray engaged in work or professional activities outside of school.} Returns event \emph{Teacher Reunion and Cultural Gift Exchange} (score\,=\,0.48). & The raw payload is rendered as a structured card listing the focus event, its temporal-causal neighbors (\emph{Arrival and Meeting Ray at Taiwan Airport}, \emph{Basketball Court Challenges}), and edge labels. A task ledger snapshot with two subtasks (t1: identify work-related visual clues; t2: determine the associated industry) is prepended. & Inform \\
\addlinespace
2 & \texttt{Search: Ray involved in business work or professional activities outside of school.} Focus unchanged (same event, score\,=\,0.45). & The sufficiency assessor computes an information increment of $0\%$ (no new keywords beyond Round~1) and appends a low-increment hint: ``\emph{The last search added limited new information. Consider whether you have enough to answer.}'' & Verify \\
\addlinespace
3 & \texttt{Search: Ray working at a shop or store behind the counter selling items.} Returns a new event \emph{Exploring the In-School Convenience Store} (score\,=\,0.43). & The focus shifts to a different event; the sufficiency assessor resets its consecutive-low counter because the new payload contributes fresh keywords (increment\,$>$\,15\%). The structured card and updated ledger snapshot are injected normally. & Inform \\
\addlinespace
4 & \texttt{Search: Ray's family shop or Ray working at a store counter outside school.} Focus unchanged (same event, score\,=\,0.53). & Increment drops to $0\%$ (same event, same keywords); the sufficiency assessor appends a second low-increment hint. The agent decides to switch retrieval mode. & Verify \\
\addlinespace
5 & \texttt{VIDEO: Ray working behind the counter, handling transactions, or wearing a uniform at the convenience store.} Drills into VideoGraph within the focus event. & The \texttt{VIDEO} prefix triggers a \emph{video drilldown}: instead of searching the EventGraph, the harness routes the query to the EntityGraph of clips 59--61. Fine-grained memories are returned (e.g., ``\texttt{<character\_0>} walks through a convenience store\dots purchases items at the checkout counter\dots tells the cashier to keep the change''). The sufficiency assessor grants a \emph{first-drilldown immunity} (new modality layer), so the low-increment counter is not incremented despite lexical overlap with the event summary. & Inform, Constrain \\
\addlinespace
6 & \texttt{Search: Ray and his family business or Ray working in a shop outside of school.} Returns a \emph{new} event \emph{Dinner and Family Time at Ray's Home} (score\,=\,0.51). & The focus shifts again; the payload contains the decisive clue: ``They also spend time looking through Ray's\dots streaming setup.'' The sufficiency assessor records high increment ($>$\,30\%) and resets the low counter. The structured card and updated ledger snapshot are injected. & Inform \\
\addlinespace
7 & \texttt{Answer: Ray is involved in the live-streaming or content creation industry, as evidenced by the streaming setup found in his home.} & --- & --- \\
\bottomrule
\end{tabular}
\caption{Harness engineering workflow on a real M3-Bench-web (``What other industry might Ray be involved in besides attending school?''; ground truth: online content creation / live streaming). The rightmost column indicates which harness mechanism fires at each step.}
\end{table*}



\begin{figure*}[t]
\centering
\begin{tcolorbox}[
  enhanced,
  title={\textbf{EventGraph Construction Prompt}},
  colback=white,
  colframe=black,
  coltitle=white,
  colbacktitle=black,
  fonttitle=\bfseries\small,
  boxrule=0.8pt,
  left=4pt, right=4pt, top=4pt, bottom=4pt,
  fontupper=\scriptsize
]

\textbf{\textit{[System Prompt]}}\\
You are a video narrative analyst. Your job is to consolidate fragmented clip-level memories into higher-level temporal and causal structures that can be used for retrieval and reasoning.\\[3pt]
\textbf{INPUT}~~You will receive: (1) PREVIOUS\_GRAPH: an existing segment-level graph built from earlier clips (may be empty on the first call). (2) NEW\_MEMORIES: a list of memories extracted every \textasciitilde30s from a long video. Each item is either: \texttt{[episodic] clip\_id=<int> | <list of strings describing visible actions / spoken content>} or \texttt{[semantic] clip\_id=<int> | <list of strings describing high-level meanings / themes>}.\\[3pt]
\textbf{GOAL}~~Maintain a compact directed graph: Nodes are higher-level SEGMENTS (each segment groups multiple clips). Edges connect SEGMENTS with explicit TEMPORAL and CAUSAL relations. Each segment stores: segment\_label, summary, clip\_ids. Each edge stores: from, to, relation\_type, relation, evidence\_clips, explicitness, confidence, rationale. This is a sliding-window refinement task: use BOTH the new memories AND the existing graph. Long-range dependencies are allowed and encouraged; temporal/causal edges do NOT need to connect adjacent segments.\\[3pt]
\textbf{PATCH-ONLY OUTPUT}~~Do NOT output the full graph. Output ONLY the changes needed to update PREVIOUS\_GRAPH using NEW\_MEMORIES. ``segments'' must contain ONLY: (a) newly created segments, and/or (b) existing segments that you MODIFY (typically by appending new clip\_ids and updating the summary). ``edges'' must contain ONLY: (a) newly created edges, (b) existing edges that you MODIFY (updated rationale/explicitness/confidence), and/or (c) edges you want to REMOVE (with \texttt{"remove": true}). If nothing changes, output \{``segments'': [], ``edges'': []\}.\\[3pt]
\textbf{INITIALIZATION}~~If PREVIOUS\_GRAPH is empty, initialize the graph from NEW\_MEMORIES. The output is still in patch format (the patch becomes the initial graph). Otherwise, update the existing graph with minimal changes: (1) assign every new clip\_id to a segment (extend or create); (2) update summaries ONLY for segments that receive new clip\_ids; (3) add temporal/causal edges supported by new evidence; (4) revise or remove old edges only if contradicted by new evidence.\\[3pt]
\textbf{SEGMENTATION RULES}~~Group adjacent clips sharing the same topic/scene/goal. Prefer 3--8 segments per batch. Each label $\leq$ 8 words, unique, natural language. Reuse and extend existing segments; do not create duplicates. Max 40 clip\_ids per segment.\\[3pt]
\textbf{SUMMARY}~~For each segment in the patch, write a detailed summary (4--8 sentences): (1) what happens (major actions/events); (2) who is involved and what they do; (3) high-level description (not step-by-step). No invented information. Do NOT rewrite summaries for segments you are not modifying.\\[3pt]
\textbf{EDGE RULES}~~Temporal edges: relation $\in$ \{``before'', ``overlap'', ``concurrent'', ``contains''\}. ``before'': from-segment happens before to-segment (do NOT use ``after''; reverse the direction instead). ``overlap'': partial time overlap. ``concurrent'': roughly simultaneous. ``contains'': from-segment temporally contains to-segment.
Causal edges: relation $\in$ \{``causes'', ``enables'', ``requires'', ``prevents''\}. ``causes'': direct causation. ``enables'': makes possible. ``requires'': precondition. ``prevents'': blocks occurrence. Explicitness: ``explicit'' if directly stated/shown, otherwise ``inferred''. Confidence: a number in [0,1]. Rationale: 1 short sentence grounded in the memories. Long-range edges encouraged. Do not assume adjacency implies causality. Prefer fewer, stronger edges.\\[3pt]
\textbf{EDGE REVISION}~~Keep old edges unchanged if still supported (do not output them). If new memories strengthen an inferred edge, output a modified version with updated evidence/rationale/confidence. If new memories weaken an edge, remove it.\\[3pt]
\textbf{EDGE REMOVAL}~~To remove an existing edge, output an edge object with the same identifiers (from, to, relation\_type, relation) and set \texttt{"remove": true}.\\[3pt]
\textbf{NON-AGGRESSIVE EDITING}~~Do not modify segments without new clip\_ids. Prefer extending over duplicating. Merge/split only when clearly necessary. Maintain sufficient granularity; do not merge distinct events due to over-abstraction.\\[3pt]
\textbf{COVERAGE}~~Every clip\_id in NEW\_MEMORIES must appear in some segment after the update.\\[3pt]
\textbf{CLIP EVIDENCE}~~Every segment must list all clip\_ids it covers (including previously covered ones if updated). Every edge must include evidence\_clips that justify the relation. Each segment should cover no more than 40 clip\_ids.\\[3pt]
\textbf{CONSTRAINTS}~~Use ONLY the provided memories; do not invent entities/events/places/dates. Output strict JSON only (no markdown, no extra text). Segment references in edges must match segment labels exactly.\\[3pt]
\textbf{OUTPUT SCHEMA (STRICT JSON)}\\
\texttt{\{"segments": [\{"segment\_label": "...", "summary": "...", "clip\_ids": [...]\}],}\\
\texttt{~"edges": [\{"from": "...(segment\_label)", "to": "...(segment\_label)",}\\
\texttt{~~"relation\_type": "temporal|causal", \\ "relation": "before|overlap|concurrent|contains|causes|enables|requires|prevents",}\\
\texttt{~~"evidence\_clips": [...], "explicitness": "explicit|inferred", "confidence": 0.0,}\\
\texttt{~~"rationale": "...", "remove": false\}]\}}\\[3pt]
\textbf{\textit{[User Template]}}\\
\texttt{PREVIOUS\_GRAPH}~~\{graph\}\\
\texttt{MEMORIES}~~\{memory\}

\end{tcolorbox}
\vspace{-10pt}
\caption{The full prompt for incremental Event Graph construction.}
\label{fig:eventgraph_prompt}
\end{figure*}

\begin{figure*}[t]
\centering

\begin{tcolorbox}[
  enhanced,
  title={\textbf{Subtask Decomposition Prompt (Planner Stage)}},
  colback=white,
  colframe=black,
  coltitle=white,
  colbacktitle=black,
  fonttitle=\bfseries\small,
  boxrule=0.8pt,
  left=4pt, right=4pt, top=4pt, bottom=4pt,
  fontupper=\scriptsize
]

\textbf{\textit{[System Prompt]}}\\
You are a task planner. The user will answer a question about a video by searching a memory bank (EventGraph + VideoGraph). Before search begins, decompose the question into atomic sub-questions to guide retrieval.\\[3pt]

\textbf{INPUT}~~Question: \{question\}; Options: \{options\} (for MCQ) or omitted (for open-ended).\\[3pt]

\textbf{DECOMPOSITION GUIDELINES}\\
1. Each sub-question must be ATOMIC --- ask ONE fact, one entity, one relation, one event.\\
2. Use stable ids \texttt{t1, t2, t3, ...} in generation order.\\
3. If a later sub-question depends on the answer of an earlier one, declare this via \texttt{depends\_on}.\\
4. Sub-questions should collectively be sufficient to answer the ORIGINAL question --- no gaps, no overlap.\\[3pt]

\textbf{QUESTION-TYPE-SPECIFIC RULES}\\
A) YES/NO or TRUE/FALSE: Decompose into 2--3 subtasks (supporting evidence, contradicting evidence, optional temporal scope).\\
B) ``Which/What'' MCQ with multiple options: One subtask per option that needs verification, or subtasks that gather discriminating facts.\\
C) MULTI-DETAIL questions (``What are the drinks for three people?''): One subtask per detail/entity mentioned.\\
D) REASON/CAUSE questions (``Why does X ...?''): Two subtasks --- the observable behavior (what) and the context/prior event explaining the cause (why).\\
E) TRULY ATOMIC (single-entity, single-attribute): Output exactly 1 subtask. This is the ONLY case where a single subtask is acceptable.\\
F) All other: Default to 2--3 subtasks that break the question into independent retrieval targets.\\[3pt]

\textbf{OUTPUT SCHEMA}~~\texttt{\{"subtasks": [\{"id": "t1", "question": "<atomic sub-question>", "depends\_on": []\}, ...], "rationale": "..."\}}

\end{tcolorbox}

\vspace{4pt}

\begin{tcolorbox}[
  enhanced,
  title={\textbf{Incremental Replan Prompt (Triggered on Stall)}},
  colback=white,
  colframe=black,
  coltitle=white,
  colbacktitle=black,
  fonttitle=\bfseries\small,
  boxrule=0.8pt,
  left=4pt, right=4pt, top=4pt, bottom=4pt,
  fontupper=\scriptsize
]

\textbf{\textit{[System Prompt]}}\\
You are the outer-loop planner for an ongoing multi-round video-QA task. The inner loop is stuck and needs you to repair the task plan --- NOT rewrite it from scratch.\\[3pt]

\textbf{INPUT}~~Original question: \{question\}; Current TaskLedger state (id / status / best\_answer / last evidence): \{ledger\_snapshot\}; Why replan was triggered: \{reasons\}.\\[3pt]

\textbf{RULES}\\
1. Preserve already-resolved subtasks verbatim. Do NOT re-add or rewrite them.\\
2. Only modify the plan when evidence has revealed the original assumption is wrong or unreachable. If the plan looks fine and we are just querying badly, prefer empty edits.\\
3. Prefer minimal edits: abandon a sub-branch, merge redundant siblings, or rewrite one subtask's text.\\
4. Any new subtask id MUST NOT collide with existing ones. Use \texttt{t\{N+1\}} style.\\
5. \texttt{update\_subtasks[*].new\_question} should be DECLARATIVE (matchable to an event summary).\\[3pt]

\textbf{OUTPUT SCHEMA}\\
\texttt{\{"update\_subtasks": [\{"id": "t2", "new\_question": "...", "reason": "..."\}],}\\
\texttt{~"abandon\_subtasks": [\{"id": "t3", "reason": "..."\}],}\\
\texttt{~"new\_subtasks": [\{"id": "t5", "question": "...", "depends\_on": []\}],}\\
\texttt{~"rationale": "<one short sentence>"\}}\\[2pt]
If no edit is warranted, return empty arrays and set rationale to ``no-op''.

\end{tcolorbox}

\caption{Task planning prompts.}
\label{fig:planning_prompts}
\end{figure*}

\begin{figure*}[t]
\centering

\begin{tcolorbox}[
  enhanced,
  title={\textbf{Multi-Round Retrieval System Prompt}},
  colback=white,
  colframe=black,
  coltitle=white,
  colbacktitle=black,
  fonttitle=\bfseries\small,
  boxrule=0.8pt,
  left=4pt, right=4pt, top=4pt, bottom=4pt,
  fontupper=\scriptsize
]

\textbf{\textit{[System Prompt]}}\\
You are given a question and some relevant knowledge. Your task is to reason about whether the provided knowledge is sufficient to answer the question. If sufficient, output [Answer]. If not, output [Search] and generate a query to retrieve additional information from a memory bank.\\[3pt]

\textbf{OUTPUT FORMAT}~~\texttt{Reason: \{reason\}; Action: [Answer] or [Search]; Content: \{content\}}\\[3pt]

\textbf{SEARCH MODES} (Content must follow one of these patterns)\\[2pt]
(1) \textbf{EVENT} (plain query): \texttt{Content: <query>} --- searches the EventGraph for the most similar event segment node; returns the focus node plus its 1-hop temporal-causal neighbors and edges.\\[2pt]
(2) \textbf{VIDEO}: \texttt{Content: VIDEO: <query>} --- retrieves fine-grained entity-level memories (actions, dialogues, attributes) from the VideoGraph within the current focus event's clips.\\[2pt]
(3) \textbf{NEIGHBOR}: \texttt{Content: NEIGHBOR: <segment\_label>} --- shifts focus to a related event by following EventGraph edges. The label must be copied exactly from the returned neighbor list.\\[2pt]
(4) \textbf{KEYFRAME}: \texttt{Content: KEYFRAME: <query>} --- loads representative keyframe images from the current focus event's clips for visual inspection.\\[3pt]

\textbf{DECISION GUIDELINES}\\
- Start with a plain query when no focus event exists.\\
- After retrieving an EVENT node, answer only if the summary directly distinguishes the correct answer.\\
- If the event is relevant but lacks specific details (name, dialogue, count, attribute), use \texttt{VIDEO:} before answering.\\
- Use \texttt{NEIGHBOR:} when the current event is off-target or adjacent events are more relevant.\\
- Use \texttt{KEYFRAME:} when remaining uncertainty is visual (color, layout, clothing, object state).\\
- Do not repeat near-duplicate queries. Switch modes or add more predicates.\\[3pt]

\textbf{QUERY WRITING}~~Shape: \texttt{<subject> + <predicate/action/attribute> + <optional context>}. A single token or bare interrogative is NOT acceptable. NEVER copy a subtask's interrogative text verbatim; TRANSFORM it into a declarative scene description.\\[3pt]

\textbf{LEDGER AWARENESS}~~The system automatically tracks evidence and updates subtask status. Each round, the system injects: (1) the current TaskLedger snapshot; (2) the scheduler-selected focus subtask; (3) optional strategy hints.

\end{tcolorbox}

\vspace{4pt}

\begin{tcolorbox}[
  enhanced,
  title={\textbf{Final Synthesis Prompt}},
  colback=white,
  colframe=black,
  coltitle=white,
  colbacktitle=black,
  fonttitle=\bfseries\small,
  boxrule=0.8pt,
  left=4pt, right=4pt, top=4pt, bottom=4pt,
  fontupper=\scriptsize
]

\textbf{\textit{[System Prompt]}}\\
You are given a question, the final state of a TaskLedger, and ALL raw retrieved knowledge from previous search rounds.\\[3pt]

\textbf{TWO SOURCES OF INFORMATION}\\
1. \textbf{TaskLedger Final State}: structured sub-question decomposition with best answers and top evidence snippets (may be incomplete if evidence binding failed).\\
2. \textbf{All Retrieved Knowledge}: the COMPLETE raw retrieval results from every search round, containing full event summaries, videograph details, and keyframe descriptions. This is the ground-truth information source.\\[3pt]

\textbf{IMPORTANT}: The TaskLedger is a summary that may have lost details. When the ledger says ``no evidence collected'' or provides only partial answers, ALWAYS check the raw retrieved knowledge for relevant information that was retrieved but not properly bound to subtasks.\\[3pt]

\textbf{EVIDENCE UTILISATION RULES (CRITICAL)}\\
- Scan BOTH the ledger AND the raw retrieved knowledge. The raw retrieval often contains answers that the ledger missed.\\
- If ANY event summary, videograph hit, or keyframe description contains an entity that could plausibly answer the question, use THAT entity as the answer.\\
- When multiple candidates appear, answer with the most contextually relevant one.\\
- For yes/no questions, answer ``Yes'' if ANY positive evidence exists; answer ``No'' only when evidence explicitly contradicts it.\\
- Do NOT reply with ``no evidence found'' when the raw retrieval contains relevant information.\\[3pt]

\textbf{MCQ DECISION RULES}\\
- Compare ALL options before choosing. Use BOTH the ledger summary AND the raw retrieved knowledge.\\
- Prefer direct evidence from raw retrieval quotes and event summaries.\\
- MUST pick exactly one option with the strongest direct support after eliminating distractors.\\[3pt]

\textbf{OUTPUT}~~\texttt{Reason: <brief option comparison citing specific evidence>; Action: [Answer]; Content: <option letter and content>}\\[2pt]
Do NOT output [Search]. Do NOT output any LedgerOps. Use real character names.

\end{tcolorbox}

\caption{Retrieval and synthesis prompts.}
\label{fig:retrieval_synthesis_prompts}
\end{figure*}

\end{document}